\documentclass[letterpaper, 10 pt, conference]{ieeeconf}
\IEEEoverridecommandlockouts
\usepackage{cite}
\usepackage{amsmath,amssymb,amsfonts}
\usepackage{algorithmic}
\usepackage{graphicx}
\usepackage{textcomp}
\usepackage{xcolor}
\usepackage{comment}
\usepackage{physics}
\usepackage{etoolbox}
\usepackage{wrapfig}
\usepackage[ruled,noend]{algorithm2e}
\usepackage{url}
\usepackage{multirow}

\newcommand{\SZ}[1]{{\textcolor{red}{SZ: #1}}}

\newcommand{\w}{w} %Weight in optimization objective
\renewcommand{\d}{d} %Determinant of inverted matrix (linear system solve)

\newcommand{\B}{\mathcal{B}} %Rigid body

\renewcommand{\t}{\mathbf{t}} %Parameters describing collision primitive
\newcommand{\p}{\mathbf{p}} %Edge point on collision primitive
\renewcommand{\v}{\mathbf{v}} %Vector describing collision primitive
\newcommand{\x}{\mathbf{x}} %State of robotic agents
\renewcommand{\r}{\mathbf{r}} %"Safety margins" for inequality constraints
\newcommand{\z}{\mathbf{z}} %Global endeffector target
\renewcommand{\b}{\mathbf{b}} %Limits for box constraints
\newcommand{\h}{\mathbf{h}} %Function for box constraints
\renewcommand{\c}{\mathbf{c}} %Local coordinates of collision primitive
\renewcommand{\l}{\mathbf{l}} %Local coordinates of end-effector

\renewcommand{\P}{\mathbf{P}} %Function that describes each point on a collision primitive

\renewcommand{\O}{\mathcal{O}} %Objective function for trajectory optimization
\newcommand{\U}{\mathcal{U}} %Objective function for collision avoidance optimization
\newcommand{\D}{\mathcal{D}} %Distance function
\newcommand{\R}{\mathcal{R}} %Regularization function
\renewcommand{\S}{\mathcal{S}} %Barrier ("softification") function
\newcommand{\K}{\mathcal{K}} %Kinematic function

\newcommand{\argmin}{\mathop{\mathrm{arg \hspace{0.1cm} min}}\limits}

\makeatletter
\patchcmd{\@makecaption}
  {\scshape}
  {}
  {}
  {}
\makeatother
\title{\LARGE \bf Differentiable Collision Avoidance Using Collision Primitives}
\author{Simon Zimmermann$^{1}$, Matthias Busenhart$^{1}$, Simon Huber$^{1}$, Roi Poranne$^{2}$, Stelian Coros$^{1}$
\thanks{This work has been submitted to the IEEE for possible publication. Copyright may be transferred without notice, after which this version may no longer be accessible.}
\thanks{$^{1}$The authors are with the Department of Computer Science, ETH, Zurich, Switzerland. %
{\tt{\small} simon.zimmermann@inf.ethz.ch; busenham@student.ethz.ch; simon.huber@inf.ethz.ch; roi.poranne@inf.ethz.ch; scoros@gmail.com}}%
\thanks{$^{2} $Department of Computer Science, University of Haifa, Haifa, Israel}%
}%

\begin{document}

\maketitle
\thispagestyle{empty}
\pagestyle{empty}

%%% Abstract
\begin{abstract}

A central aspect of robotic motion planning is collision avoidance, where a multitude of different approaches are currently in use.
Optimization-based motion planning is one method, that often heavily relies on distance computations between robots and obstacles.
These computations can easily become a bottleneck, as they do not scale well with the complexity of the robots or the environment.
To improve performance, many different methods suggested to use collision primitives, i.e. simple shapes that approximate the more complex rigid bodies, and that are simpler to compute distances to and from.
However, each pair of primitives requires its own specialized code, and certain pairs are known to suffer from numerical issues.
In this paper, we propose an easy-to-use, unified treatment of a wide variety of primitives.
We formulate distance computation as a minimization problem, which we solve iteratively.
We show how to take derivatives of this minimization problem, allowing it to be seamlessly integrated into a trajectory optimization method.
Our experiments show that our method performs favourably, both in terms of timing and the quality of the trajectory.
The source code of our implementation will be released upon acceptance.

\end{abstract}

%%% Chapters
\section{Introduction}

\noindent Collision avoidance is an integral part of robotic motion planning.
%Finding a collision-free path is often not only a question of successful task execution, but also a matter of safety, especially for human co-workers.
Cluttered environments like construction sites, where many potential collisions may occur, are burdened with great computational load.
Planning paths for multiple robots requires \emph{dynamic} and flexible collision avoidance, making the problem more difficult.
Furthermore, where human collaborators are involved, robots need to plan ahead while treating the humans as unpredictable, moving obstacles.
Indeed, the increasing complexity of tasks that robots are expected to perform certainly requires an equal increase in the efficiency of motion planning algorithms.

%finding a safe and yet efficient path is more essential than ever \SZ{rephrase}.and

%In these scenarios, it is not only important that the robots do not collide with themselves or their environment, but do also not cross paths with their robotic collaborators.

\noindent Our goal in this paper is to derive a simple, yet robust and \emph{customizable} approach for collision avoiding trajectory optimization.
A common practice with this approach is to utilize distance functions between obstacles and define motion planning as a constrained optimization problem.
In that sense, the distance functions are used to penalize proximity of the robot to obstacles.
Computing the true distance to an obstacle, however, can be computationally demanding.
Instead, previous approaches opted to use approximations in the form of collision primitives.
That is, they replace the complex geometry of robots and obstacles by simpler shapes, such as spheres, that are easy to compute distances to.
However, the accuracy of the approximation depends on the number of collision primitives used.
Indeed, an oblong shape such as the links in a robot's arm might only be faithfully represented by a multiplicity of spheres, depending on their length and widths.
Follow up work suggested using alternative primitives, such as ellipsoids, capsules, boxes, and their combinations, that can better fit to different geometries, and thus reduce the number of primitives necessary \cite{rakita2021}.
However, this typically requires specialized code that computes distances for every type of primitive pair, which could be cumbersome to maintain, especially when using gradient-based methods.
In addition, some of the computations involved are known to be numerically sensitive \cite{Eberly2015RobustCO}, which can cause numerical issues throughout the planning process, especially when derivatives are required.
%In addition, some of the computations involved, for example, the distance between two line segments, are known to be numerically sensitive when the two segments are almost parallel.%, can lead to severe convergence issues.

\noindent Our approach is to formulate distance computation as a low-level, \emph{differentiable} optimization problem, where we explicitly handle numerical issues for all primitive pairs by adding a simple regularization term.
We combine this approach with a straight-forward parameterization of different collision primitives, resulting in a light-weight and easy-to-understand implementation, and we will release the corresponding code written in C++ upon acceptance.
We show how to take derivatives of the distance computation by leveraging sensitivity analysis.
Finally, we integrate this approach into a high-level trajectory optimization problem for collision-free multi-robot motion planning with dynamic obstacles. 
We evaluate the efficacy of our method based on a variety of simulated and real-world experiments, involving single- as well as multi-robot scenarios.

%\noindent Our approach is to formulate distance computation as a low-level, \emph{differentiable} optimization problem, which can be integrated into a high-level trajectory optimization problem.
%This enables an easy-to-use, unified framework for collision-free multi-robot motion planning with dynamic obstacles.
%We evaluate the efficacy of our method based on a variety of simulated and real-world experiments, involving single- as well as multi-robot scenarios.
%The code to compute the shortest distance and the derivatives for all primitive pairs will be released upon acceptance.

%To this end, we formulate the path planning as a straight-forward constrained optimization problem.
%Here, we introduce collision avoidance in form of inequality constraints, which involve the computation of the shortest distance between two primitives.
%\SZ{Rephrase:} Instead of using a geometric approach to derive this computation for each primitive pair, we choose an \SZ{equivalent}, but more generic formulation: We pose the shortest distance computation as a small minimization problem, where we use a unified formulation for different types of primitives.
%We circumvent numerical problems by adding a simple regularization term.
%Finally, we show how to compute derivatives of this minimization problem using sensitivity analysis.
%Consequently, the inequality constraints used for collision avoidance are continuously differentiable, which makes our method suitable to be used in gradient-based methods.

\begin{figure*}%[ht]
    \centering
    \includegraphics[width=1.0\linewidth]{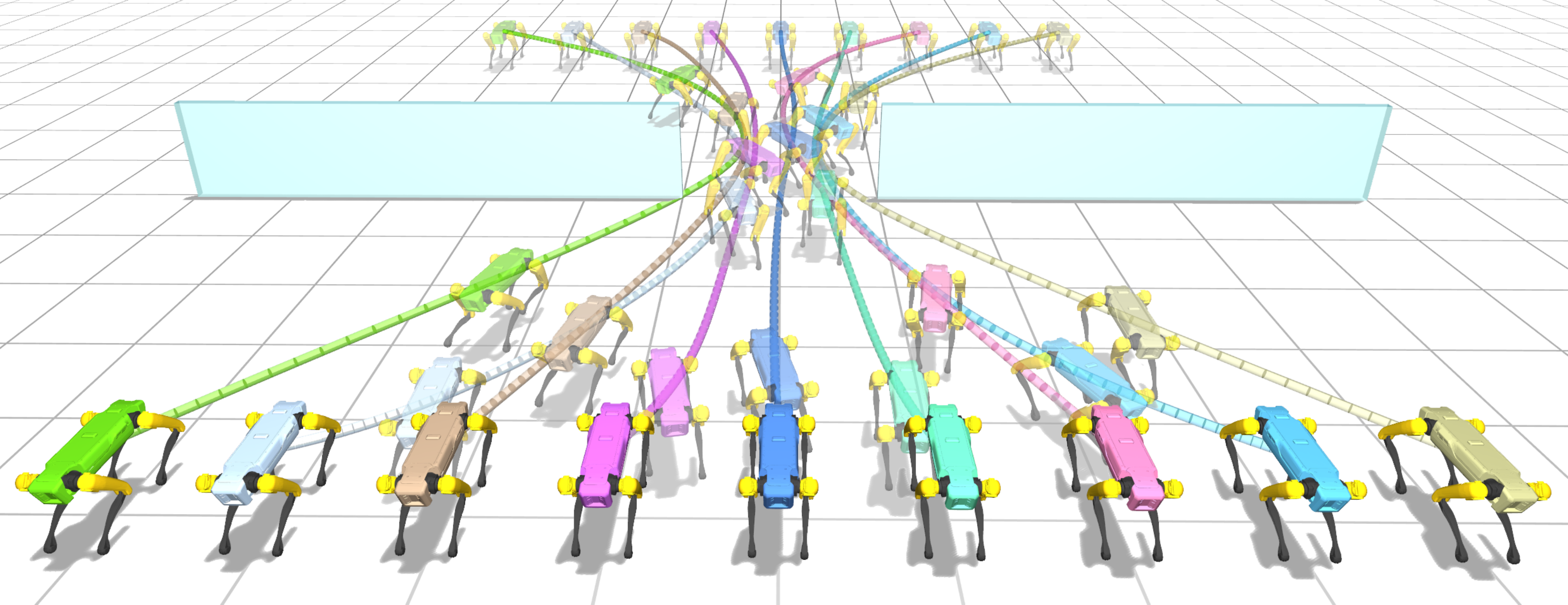}
    \caption{An armada of Spot robots are finding their way through a gap without colliding. Several poses are overlaid.}
    \label{fig:teaser}
\end{figure*}

\section{Related Work}\label{sec:related}
\noindent The literature on path planning and collision avoidance is vast, with many conceptually different approaches currently in use, each with its own benefits and drawbacks.
Therefore, we only highlight a few selected papers, and focus our attention on collision-free path planning via smooth trajectory optimization.
In these type of formulations, collision avoidance is typically incorporated in form of inequality constraints, similarly to our approach.
%In these type of formulations, the desired path or trajectory is the one that minimizes a certain objective and adheres to certain constraints.
%The objective is often a combination of several costs related to, for example, smoothness or target positions.
%The constraints, among other things, could express collision avoidance.

\noindent Some of these related methods are agnostic to the representation of the obstacles, and rely on collision \emph{detection} only to define an optimization problem \cite{alonso-mora_collision_2015}.
A straightforward approach is to describe the robot and obstacles as \emph{polyhedrons}, which leads to disjunctive linear inequalities that can be addressed in a mixed-integer problem \cite{blackmore2006optimal}.
Using Farkas' lemma, it is also possible to convert these disjunctive inequalities into standard inequalities \cite{gerdts2012path}.
A more recent approach presented in \cite{zhang_optimization-based_2020} can also handle more general ellipsoidal obstacles.
It employs duality theory to formulate simple, differentiable constraints.
The resulting optimization problem is then solved using IPOPT \cite{wachter_implementation_2006}, a generic interior point solver.

\noindent Another common approach we briefly mention uses \emph{potential fields}.
First presented in \cite{khatib_real-time_1985}, the idea is to create an artifical potential that attracts the robot to the goal and a potential that \emph{repels} the robot away from obstacles.
A few examples can be found in \cite{ge_new_2000,yngve_robust_2002,tao_path_2018,gai_6-dof_2019,safeea_collision_2020,scoccia_collision_2021}.
A related approach uses distance fields instead of potential fields.
It is used to constrain the distance between a robot and an obstacle to be greater than a certain safety margin.
Notable examples are CHOMP \cite{chomp} and STOMP \cite{stomp}, both of which sample a distance field on a regular grid as a preprocessing step.
While this approach is fairly efficient, it is limited to static obstacles and a single robot scenarios.
In addition, the memory requirements might be prohibitive.

\noindent We take an approach that, in some sense, lies between the above-mentioned methodologies.
We utilize \emph{collision primitives}, i.e. simple shapes that allow simple and efficient distance computations, to approximate the robots and obstacles in the scene.
This has been a common strategy in past as well as in recent work.
Particularly spheres are applied as collision primitives in countless papers due to their ease-of-use \cite{Duenser2018,Gaertner2021}.
Ellipsoids and capsules are also commonly found \cite{Chioniso2013,Brito2019,rakita2021,Khoury2031}, as well as boxes \cite{Toussaint2015}.
Our approach unifies all of these primitives in an easy-to-use manner.

\noindent An alternative methodology is presented by Schulman et al \cite{schulman_motion_2014}.
There, robot bodies and obstacles are represented by convex shapes, using a support mapping representation.
The distance between two shapes is formalized using signed distance functions, which are computed using the Gilbert–Johnson–Keerthi (GJK) distance algorithm.
Collision avoidance is then incorporated into a high-level sequential convex optimization problem, where they approximate the corresponding derivatives by linearizing the signed distance function.
Our approach presents a different strategy: We start with basic collision primitives like spheres and extend the concept to more complex shapes.
With this approach, we can easily handle degenerate cases that plague the GJK algorithm and result in numerical issues, which occur surprisingly often in practice.

\section{Method}
\label{sec:method}

\noindent To simplify the exposition, we first describe our approach using a simple example of two rigid bodies.
The method is easily generalized to multiple robots and obstacles, both stationary or mobile, which we discuss at the end of this section.

%------------------------------------------------------------------------
\subsection{Collision-Free Motion Planning}
\noindent Our goal is to plan a smooth, collision-free trajectory for two free-floating rigid bodies denoted by $\B^{A}$ and $\B^{B}$, respectively.
We formulate this as a \emph{time-discretized} trajectory optimization problem.
To this end, let $\x_{i}^{A}$ and $\x_{i}^{B}$ be the corresponding states of $\B^{A}$ and $\B^{B}$, comprising of a rotation and a translation in world coordinates at a specific trajectory step $i$.
We represent the entire state by stacking both states into one vector $\x_{i} = (\x_{i}^{A}, \x_{i}^{B})$.
Then, $\x := (\x_{1}, ..., \x_{N})$ represents the entire state trajectory that we want to optimize, consisting of a total number of $N$ steps.
Furthermore, let $\D^{AB}(\x_{i})$ be the squared distance between the two rigid bodies at trajectory step $i$, defined as the \emph{shortest} squared distance between any pair of points on the two rigid bodies in their respective states.
%We used the squared distance to simplify mathematical expressions later.
We include a safety margin with each of the individual rigid bodies, which we denote by $r^{A}$ and $r^{B}$, respectively.
%
%\begin{wrapfigure}[7]{rt}{0.50\linewidth}
%\centering
%\vspace{-12pt}\hspace{-15pt}\includegraphics[trim=-2pt 12pt 0pt 0pt,width=1.0\linewidth]{figures/rigidbodies.pdf}
%\end{wrapfigure}
%
%These are visualized in the inset.
Collision avoidance is formulated as an inequality constraint, which forces the shortest distance between the rigid bodies the be greater than the sum of their safety margins.
With these definitions, we write the trajectory optimization problem as
\begin{subequations}
\label{eq:rb_optimization}
\begin{alignat}{2}
& \!\!\min_{\x} & \hspace{0.2cm} & \O(\x) \label{eq:rb_objective} \\\ 
& \text{s.t.}   & & \D^{AB}(\x_{i}) \geq r^{A} + r^{B} \hspace{0.2cm} \forall i \in {1, ..., N} \label{eq:rb_inequality}
\end{alignat}
\end{subequations}
with objective $\O$, which typically consists of two terms.
The first term expresses a goal objective, which matches the state of the rigid bodies at a given trajectory step $i$ to a predefined target $\bar{\x}_{i}$.
The second term is a regularization term that encourages smooth motions by penalizing high accelerations throughout the trajectory.
We discretize the acceleration as $\Ddot{\x}_{i} \approx \frac{\x_{i} - 2\x_{i-1} + \x_{i-2}}{h^2}$, where $h$ is the step duration.
The objective (\ref{eq:rb_objective}) is then
\begin{equation}
    \O(\x) = \sum_{i\in I}\norm{\x_i - \bar{\x}_i}^2 + \w_{S}  \sum_{i = 1}^{N}\norm{\Ddot{\x}_{i}}^2,
\end{equation}
where $I$ is the set of fixed target states $\bar{\x}_i$, and $w_{S}$ denotes the regularization weight.

%------------------------------------------------------------------------

\begin{table*}%[ht]
    \renewcommand{\arraystretch}{1.3}
    \begin{center}
        \caption{Collision primitives used in this work. All can be parameterized using Eq. (\ref{eq:primitive_point_function}). For simplicity, we temporarily drop all dependencies from $\x$.}
        \label{tab:collision_primitives}
        
        %\begin{tabular}{p{2cm} | p{3.5cm} | p{3.5cm} | p{3.5cm} | p{3.5cm}}
        \begin{tabular}{c | c | c | c | c}
            \textbf{Primitive} & Sphere & Capsule & Rectangle & Box\\
            \hline \hline
            $\P(\x, \t) = $ & $\p$ & $\p + t_1 \v_{1}$ & $\p + t_1 \v_{1} + t_2 \v_{2}$ & $\p + t_1 \v_{1} + t_2 \v_{2} + t_3 \v_{3}$\\
            \hline
            & \includegraphics[width=1.15cm]{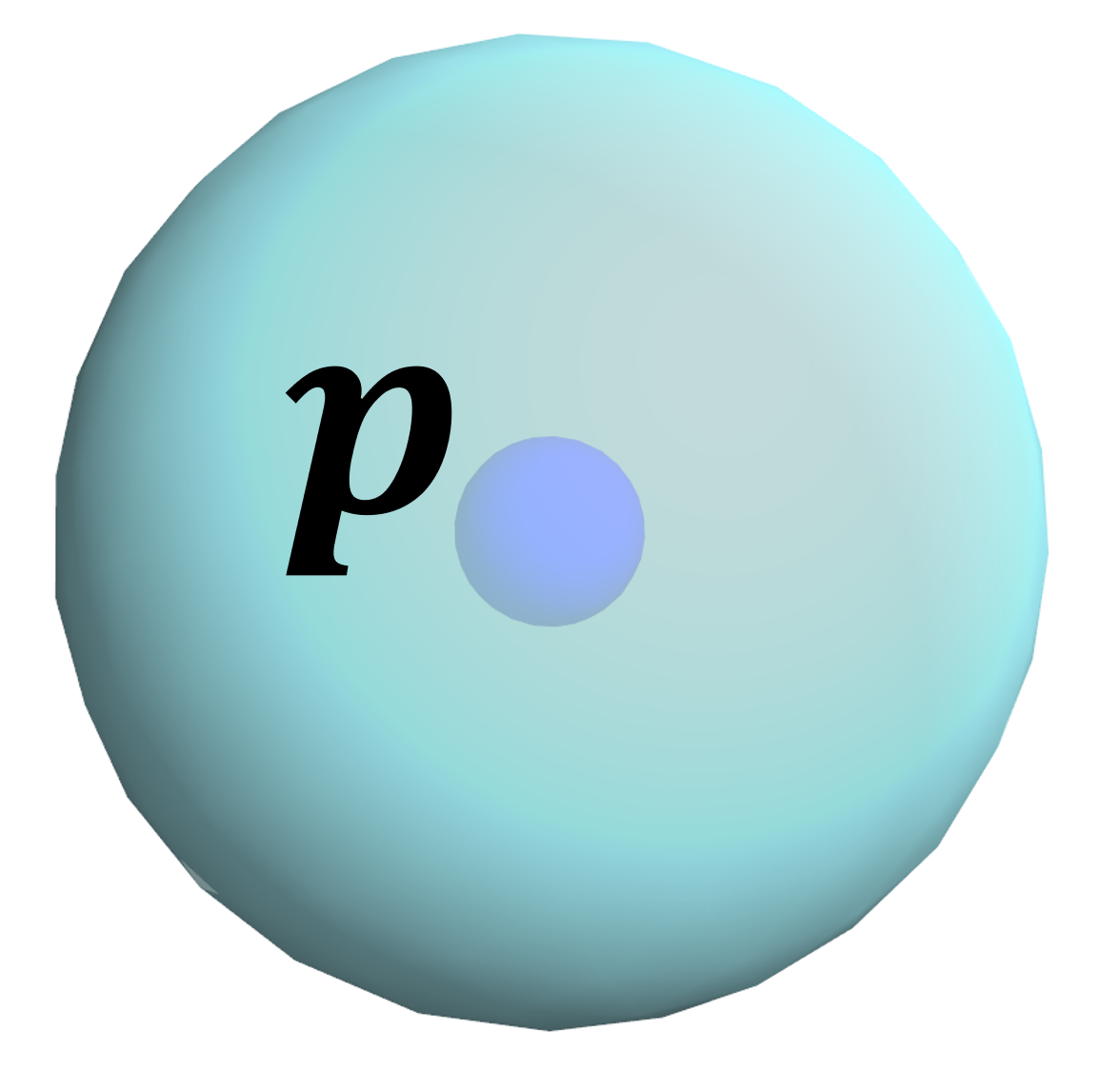} & \includegraphics[width=1.9cm]{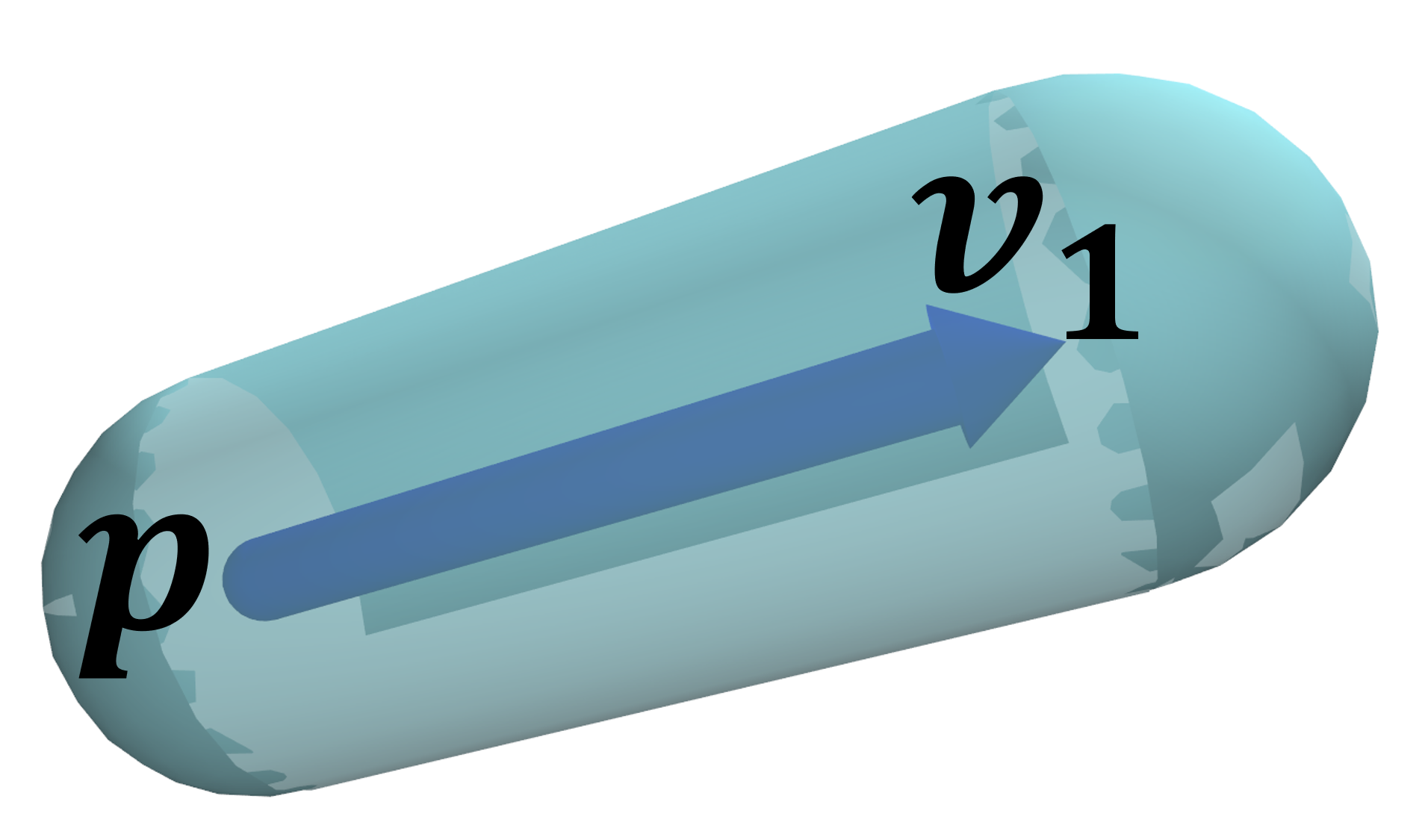} & \includegraphics[width=1.75cm]{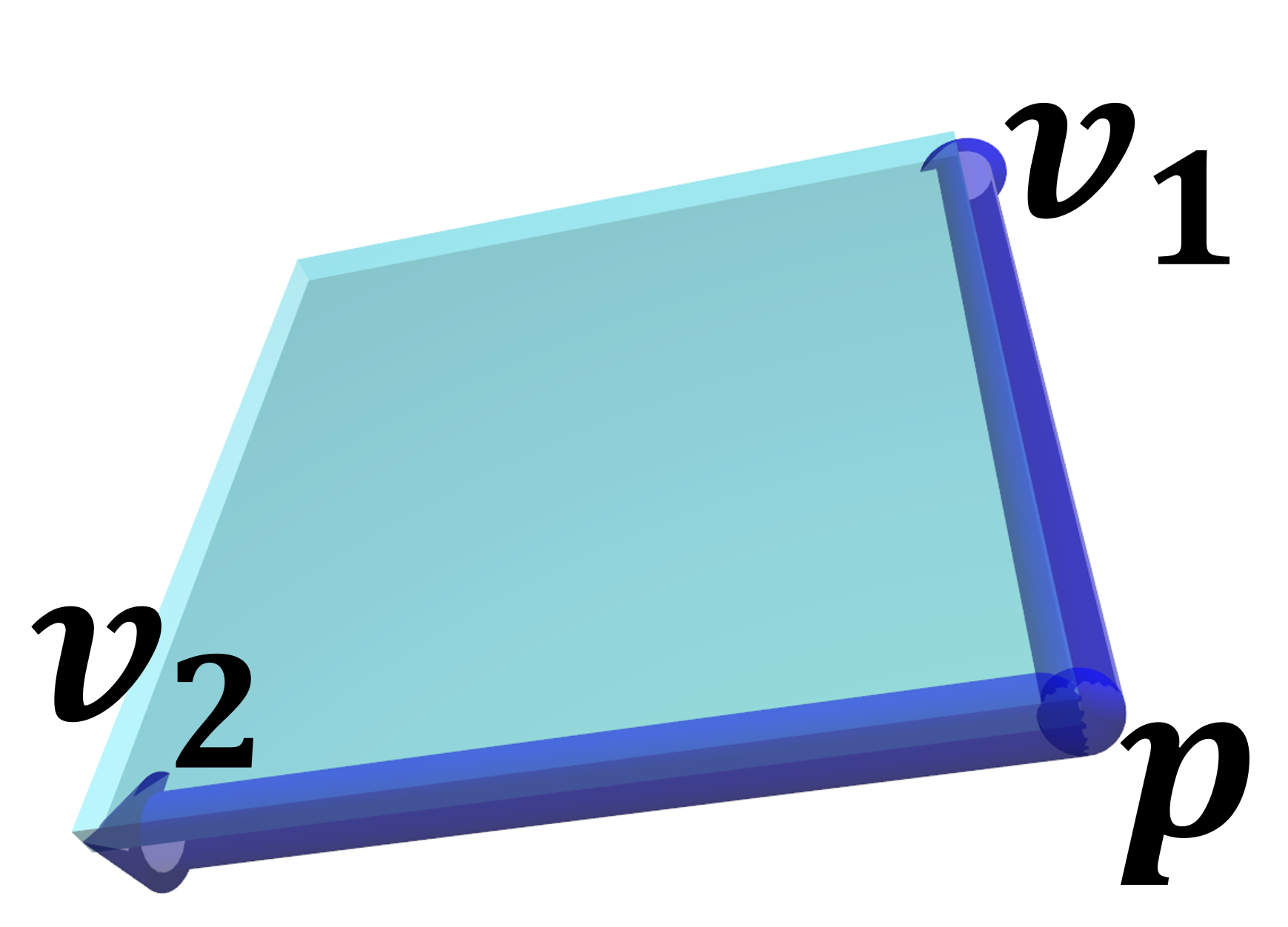} & \includegraphics[width=1.75cm]{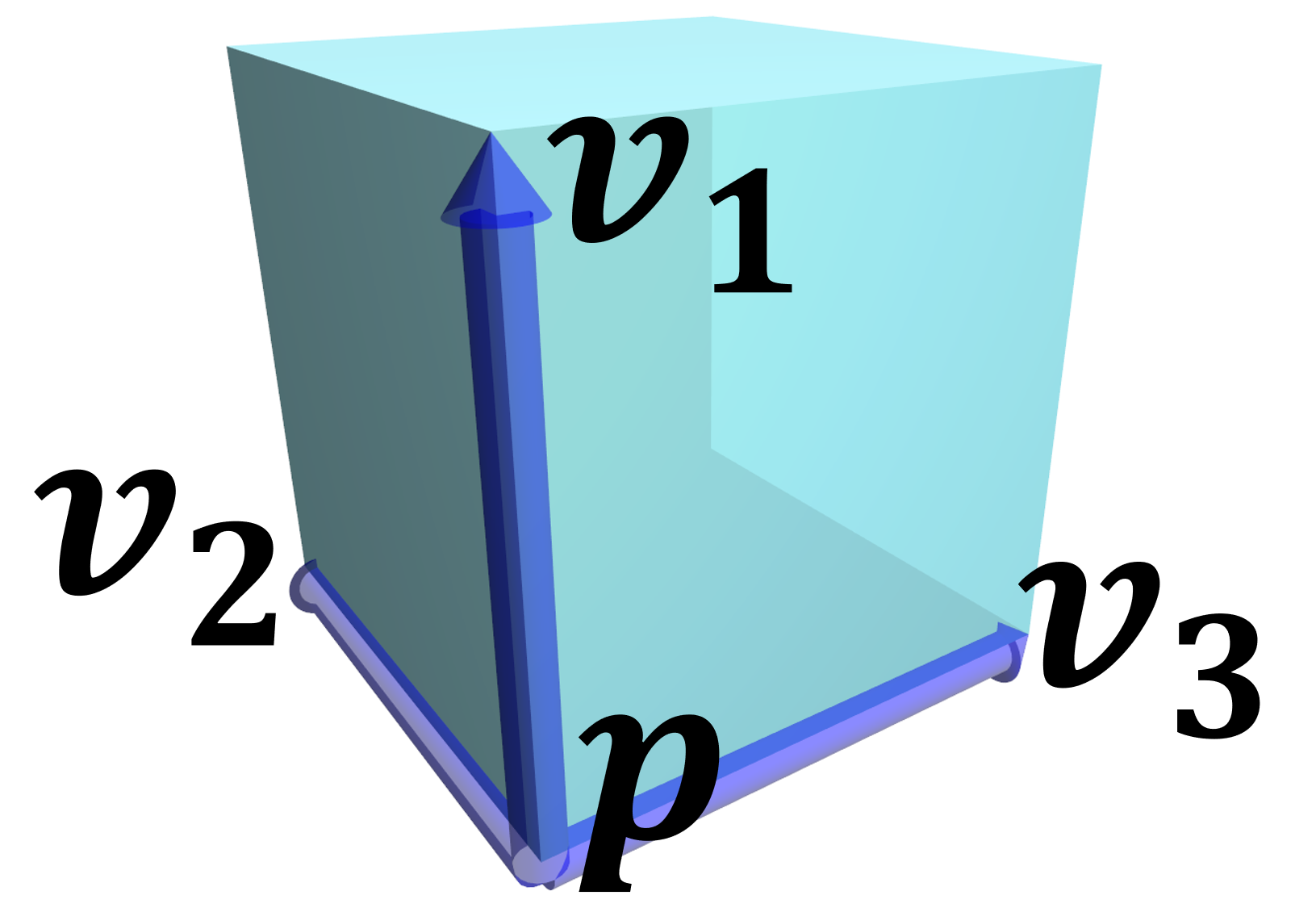} \\
        \end{tabular}
    \end{center}
    \renewcommand{\arraystretch}{1.0}
\end{table*}

\subsection{Collision Primitives}
\noindent The problem stated in Eq. (\ref{eq:rb_optimization}) is an idealized one, as $\D^{AB}$ is set to compute the \emph{true} distance between the two bodies $\B^A$ and $\B^B$.
However, this is often not practical since the computation of this distance -- and its derivatives -- can be expensive for general shapes.
Instead, a common practice is to approximate rigid bodies using simple collision primitives, which are endowed with simpler distance computation routines.
The simplest primitives used (for trajectory optimization) are perhaps spheres. 
They can be seen as points with a safety margin corresponding to the radius, and therefore require only point-to-point distances.
Furthermore, when a rigid body has a shape that can not be approximated by one sphere sufficiently well, several smaller spheres can easily be used together to obtain a better approximation.
The drawback, however, is that some shapes require \emph{too} many spheres to be approximated well, offsetting the computational benefit of using them in the first place.
In these cases, more elaborate primitive shapes such as ellipsoids, capsules, or boxes can be used at the cost of slightly increased computation times, as mentioned in Sec.~\ref{sec:related}.
Therefore, creating an efficient approximation of an object using collision primitives requires balancing the complexity of the primitives with their numbers.
To simplify the process, we propose a unified formulation for a set of different collision primitives.
This formulation is based on the simple observation that common primitives can be described \emph{parametrically} using a point $\p$ and a varying number of vectors $\v_l$, each scaled by a parameter $t_l$.
Hereby, the number of vectors and scale parameters depends on the type of primitive.
For example, a primitive with no scaled vector simply results in a point $\p$, which, combined with its radius as a safety margin, results in a sphere collision primitive.
In the same manner, a point and one scaled vector $\p + t_1 \v_1, \hspace{0.1cm} 0\leq t_1 \leq 1$ describes a line segment, which can be used as a capsule primitive by choosing its radius as a safety margin again.

\noindent We now extend this formulation for the general case: For a rigid body $\B$ with state $\x_{i}$, let $\t_i$ be the set of parameters stacked into one vector that describe all points lying on the corresponding collision primitive.
In the following, we neglect the subscript $i$ for brevity, and consequently use $\x$ to represent the state of a single trajectory step.
Let $\P(\x, \t)$ be a function that returns a specific point on the primitive as a function of $\x$ and $\t$.
In this work, we select a set of collision primitives that can all be described by
\begin{equation}
\label{eq:primitive_point_function}
    \P(\x, \t) = \p(\x) + \sum_{l = 1}^{L} t_l \v_l(\x), \quad 0\leq t_l \leq 1, \hspace{0.1cm} \forall l
\end{equation}
where $\p(\x)$ is a point on the primitive, and $\v_l(\x)$ denote a varying number $L$ of vectors, depending on the type of primitive.
All primitive shapes used in this work are visualized in Table \ref{tab:collision_primitives}.
We find that this set of simple primitives is sufficient for our application.
Nevertheless, we note that more constraints can be added to create different shapes, e.g. $\sum_{l=1}^{L} t_l \leq 1$ for a simplex.
Additionally, any convex collision primitive that can be described parametrically in a continuously differentiable manner with respect to both $\x$ and $\t$ can be used as well.

\subsection{Shortest Distance Computation}
\noindent We return to the two rigid bodies $\B^{A}$ and $\B^{B}$, and assume their collision primitives are given by $\P^{A}(\x^{A}, \t^{A})$ and $\P^{B}(\x^{B}, \t^{B})$, where we stack $\t = (\t^{A}, \t^{B})$.
%In order to describe the distance computation between two primitives, we come back to the example of the two rigid bodies as described in Eq. (\ref{eq:rb_optimization}): 
%Let $\P^{A}(\x_{i}^{A}, \t^{A})$ and $\P^{B}(\x_{i}^{B}, \t^{B})$ describe the collision primitives that approximate the shape of the rigid bodies $\B^{A}$ and $\B^{B}$, respectively.
%Recall that $\x_{i} = (\x_{i}^{A}, \x_{i}^{B})$ denotes the states of both rigid bodies stacked into one vector.
%For simplicity, we apply the same notation for $\t = (\t^{A}, \t^{B})$.
The squared shortest distance between the two primitives can be written as
\begin{equation}
\label{eq:distance_function}
    \D^{AB}(\x) = \min _{0\leq\t\leq 1}\D^{AB}(\x,\t)
\end{equation}
with 
\begin{equation}
    \D^{AB}(\x,\t) = \| \P^{A}(\x^{A}, \t^{A}) - \P^{B}(\x^{B}, \t^{B}) \|^2.
\end{equation}
This problem can be solved analytically for the cases we discuss in this paper.
However, this must be done in a case-by-case manner, and additionally can result in some numerical issues, especially when computing derivatives.
For example, for two capsule primitives parameterized as line segments, these issues occur when the two lines are close to being parallel, as the closest points between them go to infinity \cite{Eberly2015RobustCO}.
Similar issues arise for other pairs of primitives as well, and we find that they frequently occur in practice.
Instead of treating each individual case, we propose a more generic approach: We solve Eq. \eqref{eq:distance_function} using iterative numerical optimization.
This also allows us to include regularization terms, and to make the collision constraints \emph{soft}, which in turn makes collision avoidance more robust. % and permits penetrations in a controllable manner.
Concretely, we add a regularization term
\begin{equation}
    \R(\t) = \|\t - 0.5\|^2,
\end{equation}
which regularizes the $\t$'s such that the resulting points on the primitives are closer to their center.
This simple term effectively avoids the numerical issues described above.
In addition, we replace the box constraints $0\leq\t\leq 1$ in Eq. \eqref{eq:distance_function} by soft barrier constraints.
Specifically, we define two unilateral quadratic barrier functions as
\begin{equation}
    \label{eq:barrier_functions}
    \S_{l}^{+}(t) = 
    \begin{cases}
     0 & t \leq l \\
     (t-l)^2 & t > l
    \end{cases}, \quad
    \S_{l}^{-}(t) = 
    \begin{cases}
     (t-l)^2 & t \leq l \\
     0 & t > l
     \end{cases}.
\end{equation}
Thus, we rewrite the optimization problem (\ref{eq:distance_function}) as
\begin{equation}
    \label{eq:CA_soft_opt_problem}
    \U^{AB}(\x) = \min_{\t} \hspace{0.1cm} \U^{AB}(\x, \t)
\end{equation}
with
\begin{equation}
    \U^{AB}(\x, \t) = \D^{AB}(\x, \t) + \w_{R} \R(\t) + \w_{C}(\S_{1}^{+}(\t) + \S_{0}^{-}(\t)),
\end{equation}
where $\w_R,\w_C$ are the regularization and penalty weights.
We solve this unconstrained optimization problem using Newton's method.

\subsection{Computing Derivatives}
\noindent To solve the motion planning problem (\ref{eq:rb_optimization}), we want to be able to compute the derivatives of the distance function $\D$ w.r.t. to $\x$. 
Following the derivation above, $\D$ is not only a function of $\x$, but also of $\t$, which in turn has to be seen as a function of $\x$ as well.
Consequently, $\D(\x) := \D(\x, \t(\x))$, and computing the gradient requires the use of the chain rule
\begin{equation}
    \label{eq:gradientDistanceFunction}
    \dv{\D^{AB}}{\x} = \pdv{\D^{AB}}{\x} + \pdv{\D^{AB}}{\t} \dv{\t}{\x}.
\end{equation}
The Jacobian $\dv{\t}{\x}$ is known as the \emph{sensitivity matrix}, as it describes how $\t$ changes with respect to changes in $\x$.
Since $\t$ is computed numerically by solving the optimization problem (\ref{eq:CA_soft_opt_problem}), there is no direct analytical expression to compute $\dv{\t}{\x}$.
However, as described in \cite{zimmermann2019optimal}, we can readily compute this Jacobian by leveraging the implicit function theorem.
For our particular application, it can be applied when the gradient of the optimization problem (\ref{eq:CA_soft_opt_problem}) is zero, i.e. $\pdv{\U^{AB}}{\t} = 0$.
This is the case when the optimization has been solved.
Then, by using the resulting solution for $\t$, the sensitivity matrix can be computed analytically as
\begin{equation}
    \dv{\t}{\x} = -\pdv[2]{\U^{AB}}{\t}^{-1}\pdv[2]{\U^{AB}}{\t}{\x}.
\end{equation}
We refer the reader to \cite{zimmermann2019optimal} for a full derivation, and simply state the Hessian for our particular use-case.
Just as in \cite{zimmermann2019optimal}, we use an approximation of the true Hessian, as it avoids the costly computation the higher-order terms.
This results in the expression
\begin{equation}
    \label{eq:hessianDistanceFunction}
    \dv[2]{\D^{AB}}{\x} \approx \left( \dv{\t}{\x}^T \pdv[2]{\D^{AB}}{\t} + 2 \pdv[2]{\D^{AB}}{\x}{\t} \right) \dv{\t}{\x} + \pdv[2]{\D^{AB}}{\x},
\end{equation}
where we can re-use the Jacobian from the gradient computation.

\subsection{Extension to Robot Motion Planning}
\label{subsec:multiRobotMotionPlanning}

\begin{comment}
    \textbf{The state} of a robot can be fully described by a vector $\x \in \mathbb{R}^{n + 6}$ with $n$ being the number of hinge joints and 6 for the position and orientation of the root body. 
    Over the full trajectory we have a specific $x_i$ with $i \in \mathbb{Z}_{N}$ and $N$ the number of time steps.
    For a scene with multiple robots we stack together the state vectors $x_i = (x_i^1, ..., x_i^M)$ where $M$ is the number of robots in the scene. 
    Then multiple trajectories are described by stacking the vectors: $x = (x_1, ..., x_N)$.\\
    \textbf{Any local point} w.r.t to a rigid body of the robot can then be expressed in world coordinates at the chosen time step by starting at the root and applying the individual rotations specified in the corresponding state vector $x_i$. Let $\K(\x_i^j)$ be this forward kinematics function.\\
    \textbf{Collision Primitives} can be attached to a rigid body at any point $p_L$ on these primitives and can be translated to world coordinates through $\K$.
    This means for the derivative of the function $\D$ that there is an additional chain rule step $\pdv{\K}{\x}$ which is just the robot jacobian.\\~\\
\end{comment}

\noindent As a final step, we extend our formulation from two rigid bodies to the case of robotic motion planning.
To this end, we model a robot as a kinematic arm with floating base:
Let $\x_{i} \in \mathbb{R}^{n+6}$ denote the state of a robot at trajectory step $i \in [1, ..., N]$, where $N$ and $n$ denote the total number of trajectory steps and joint angles, respectively.
The state $\x_{i}$ consists of the position and orientation of the base, as well as the joint angles of the arm.
Same as before, we describe the entire robot trajectory by stacking all states $\x_{i}$ into one vector $\x := (\x_{1}, ..., \x_{N})$.
Using these definitions, we can now extend the optimization problem presented in Eq. (\ref{eq:rb_optimization}).
Since we are now handling kinematic arms, it is more convenient to include end-effector targets into our formulation:
Let $\z_{i}$ be the target pose in global coordinates for the robot's end-effector at a predefined trajectory step $i$.
Let $\K(\x_i, \l)$ be the forward kinematics function that transforms the local coordinates $\l$ of the end-effector into its global pose at state $\x_i$.
We can then formulate an inverse kinematics (IK) objective as
\begin{equation}
    \norm{\K(\x_i, \l) - \z_i}^2
\end{equation}
for each end-effector and trajectory step we wish to set a target for.
Furthermore, physical limitations of an individual robot like joint, velocity and acceleration limits need to be considered for motion planning.
This can be included into the formulation in form of box constraints
\begin{equation}
    \b_{l} \leq \h(\x_{i}) \leq \b_{u}, \hspace{0.1cm} \forall i
\end{equation}
where $\b_{l}$ and $\b_{u}$ denote the lower and upper bounds, respectively, and $\h(\x_{i})$ returns the corresponding value at trajectory step $i$.
Velocities and accelerations are approximated using finite differences.

\begin{figure} %[h]
    \centering
    \includegraphics[width=1.0\linewidth]{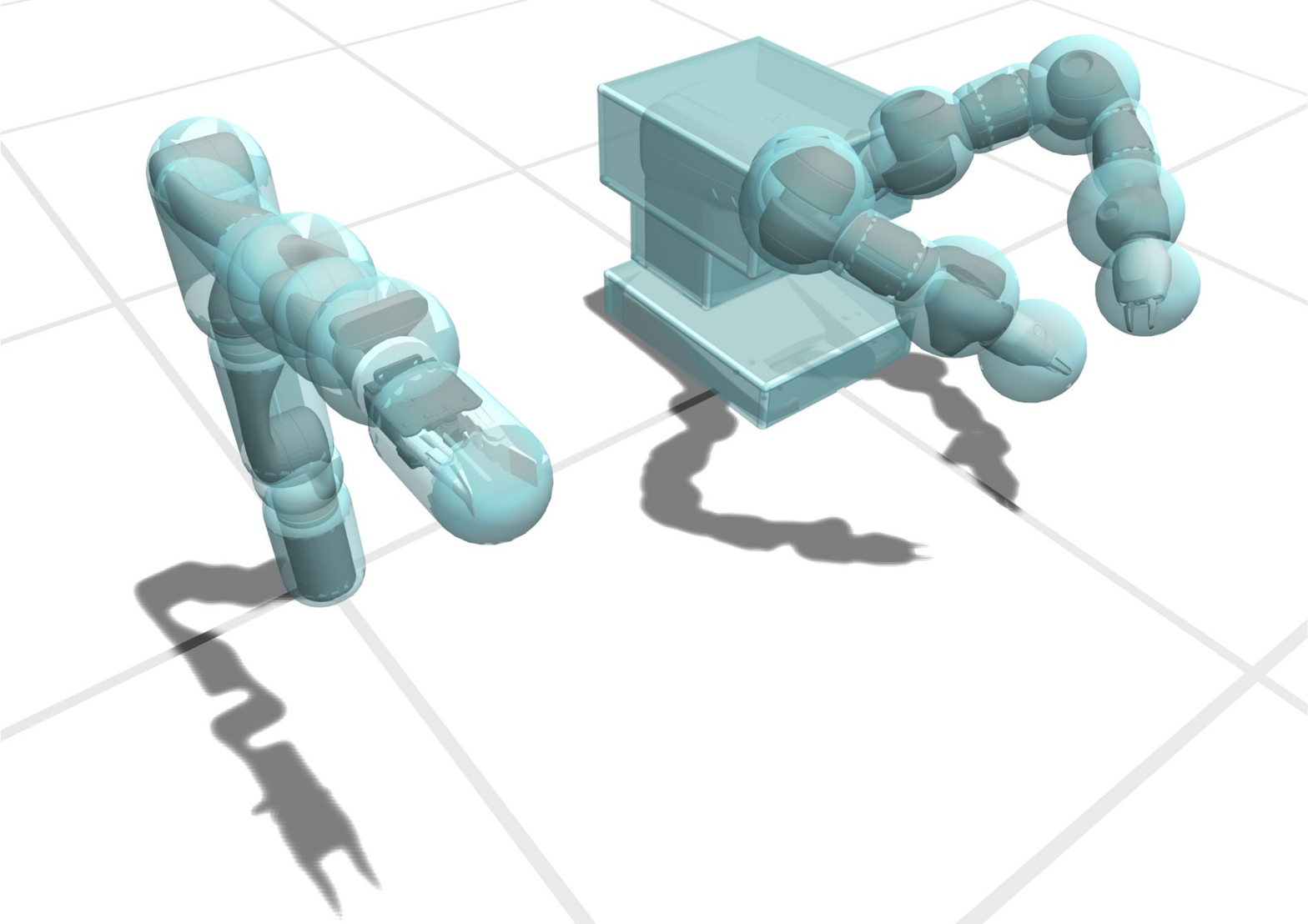}
    \caption{The YuMi (right) and Kinova (left) robot approximated by collision primitives. The YuMi uses three boxes for the body and five spheres and two capsules per arm. The Kinova uses three capsules and four spheres.}
    \label{fig:robotCollisionApprox}
\end{figure}

\noindent In order to apply collision avoidance, a robot can be approximated by different collision primitives in the same manner as in the single rigid body case.
Examples for two different robots are given in Fig. \ref{fig:robotCollisionApprox}.
This can be done conveniently by expressing a specific collision primitive in local coordinates of the body it approximates.
Since the robot's bodies are connected via a kinematic chain, the formulation as presented in (\ref{eq:rb_inequality}) needs to be slightly adjusted, as the world coordinates of the collision primitives have to be computed using the forward kinematics function $\K$.
Therefore, the inequality constraints are written as
\begin{equation}
    \label{eq:robotcollisionavoidance}
    \D^{ab}(\K(\x_i, \c^a), \K(\x_i, \c^b)) \geq \r^a + \r^b, \hspace{0.2cm}
    \forall i, \forall {a,b} \in \mathbb{C},
\end{equation}
where $\c^a$ and $\c^b$ denote the local coordinates of collision primitive $a$ and $b$, respectively, and $\mathbb{C}$ includes all collision primitives used in the scene.
These constraints formulate self-collision avoidance of an individual robot for all trajectory steps.
In case primitives are used to approximate static obstacles, the dependency on $\K$ can be dropped, as the state of the obstacle can be written directly in world coordinates.

\begin{comment}
    Now the only missing piece are the distance constraints for each robot vs robot respectively each collision primitive vs collision primitive. 
    It integrates itself neatly in the presented formulation by stating: 
    \begin{equation}
    \begin{aligned}
    \D^{ab}(\K(\x_i, \cl_a), \K(\x_i, \cl_b)) \geq \r^j + \r^k, \\
    \forall \cl_a \in \mathcal{C}^a, \forall \cl_b \in \mathcal{C}^b, \forall {a,b} \in \mathcal{R}    
    \end{aligned}
    \end{equation}
    %
    where the $\D$ is defined as in Equation \ref{eq:distance_function}, $\mathcal{C}^j$ is the set of all collision primitives of robot $j$ and $\mathcal{R}$ is the set of all robots.
    In case one of the collision primitives is a world collision obstacle i.e. does not belong to a robot, $\K$ is just the identity function. 
    For calculating the derivatives we followed the derivation in \ref{eq:deriv_distance} with one additional step in the chain-rule, the need for calculating the 
    robot jacobian, $\pdv{\K}{\x}$. 
\end{comment}

\noindent This formulation can easily be extended to plan motions for multiple robots.
To this end, we simply stack the states of all robots in the scene into one large optimization vector $\x$.
Then, the inequality constraints (\ref{eq:robotcollisionavoidance}) not only cover self-collision avoidance, but also avoidance between different robots.
We solve this multi-robot motion planning problem using Newton's method.
To this end, we convert all inequality constraints into soft constraints using barrier functions as presented in Eq. (\ref{eq:CA_soft_opt_problem}).
Then, we can compute the gradient and Hessian of the total objective $\O(\x)$ in order to apply Newton steps until convergence.
An overview of the overall solving strategy is given in Algorithm \ref{alg:newtonsMethod}.
We note that in practice, we only compute the exact distance and its derivatives between individual primitive pairs if they are in close proximity to another.
To this end, we first roughly estimate the distances between each pair according to the distance of their center points.

\begin{algorithm*}%[t]
    \caption{Collision-free multi-robot kinematic trajectory optimization using Newton's method}
    \KwIn{Total objective $\O(\x)$ (including collision avoidance constraints as soft constraints), initial $\x$}
    \KwOut{Optimal motion trajectories $\x^*$}
    \While{convergence criterion not reached} {
        %Identify primitive pairs that are in close proximity by roughly estimating their distance at each trajectory step \\
        Compute $\t$ by solving optimization problem (\ref{eq:CA_soft_opt_problem}) for each primitive pair in close proximity using Newton's method \\
        Compute gradient $\dv{\O}{\x}$ and Hessian $\dv[2]{\O}{\x}$ (includes the use of Eq. (\ref{eq:gradientDistanceFunction}) and (\ref{eq:hessianDistanceFunction}) by following the chain rule)\\
        Compute search direction $\Delta \x$ by solving linear system $\dv[2]{\O}{\x} \Delta \x = -\dv{\O}{\x}$ \\
        Run backtracking line search on $\alpha$ in $\x := \x + \alpha \Delta \x$
    }
    \label{alg:newtonsMethod}
\end{algorithm*}
\section{Results}

\noindent We evaluate the efficacy of our method on a variety of simulated and real-world experiments.
These comprise of path planning tasks for different robotic platforms in both single- and multi-robot scenarios.
Our method outputs collision-free, smooth motion trajectories for every robot in the scene.
It avoids self-collision for a single robot, collisions between two different robots, and collisions with obstacles.
We encourage the reader to watch the accompanying video, where all conducted experiments are shown.
An extended version of this video can be found online\footnote{Accompanying video: \url{https://youtu.be/et0bu--wuy4}}.
More quantitative information about the individual experiments can be found in Table \ref{tab:quantitative_results}.
An Intel Core i7-7709K 4.2Ghz PC has been used to record all measurements.

%-----------------------------------------------------------------------
\subsection{Experiments}
\noindent \textbf{Interactive Avoidance.}
We demonstrate our collision avoidance method in an interactive, real-world setting using a dual-armed YuMi robot \cite{yumi_robot}.
Two users move around obstacles that tracked by a motion capture system and approximated by collision spheres in the planning framework.
The robot successfully avoids the obstacles while ensuring that its two arms do not collide with each other.
Motion planning in this case is run in a receding horizon fashion.
The motion optimization objective for the robot contains a regularization term that encourages the robot to return to its rest pose when it is undisturbed by the obstacles.

%-----------------------------------------------------------------------

\noindent \textbf{Legged Armada.}
This experiment demonstrates kinematic path planning for the bases of multiple legged robots.
We use the simulation model of several Boston Dynamics' Spot robots \cite{spot_robot}, which are modeled as floating bases with pose and velocity constraints that resemble those of the physical Spot.
The accompanying video and Fig. \ref{fig:teaser} show different scenarios where the robots are tasked with switching positions while avoiding each other as well as some large world obstacles.
We note that motion planning is only conducted for the robot's bases, but leg motions are kinematically computed in post-processing for visualization purposes.

%-----------------------------------------------------------------------

\begin{figure} %[h]
    \centering
    \includegraphics[width=1.0\linewidth]{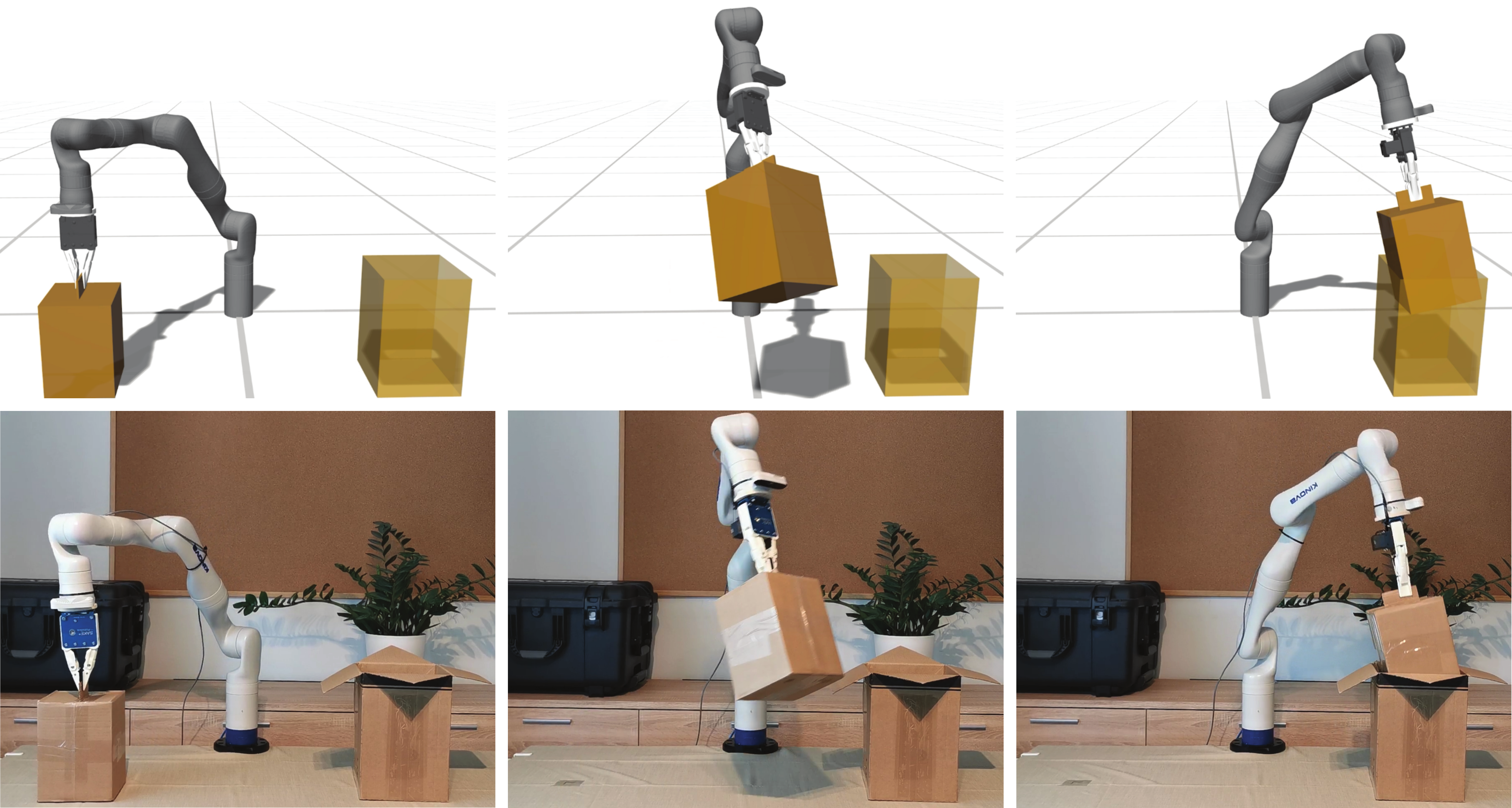}
    \caption{Three instances from the \emph{package packing} demonstration, visualized both in simulation and in real-world: the Kinova robot places a package into a slightly larger box.}
    \label{fig:packagePacking}
\end{figure}

\noindent \textbf{Package Packing.}
A statically mounted robot arm picks up a package, and places it in a slightly larger, empty box.
The empty box is approximated by four rectangular collision primitives, one for each closed side.
This experiment is also conducted in the real world by passing the nominal trajectories computed in simulation onto the physical robot in an open-loop manner.
We use a Kinova Gen3 7Dof robot \cite{kinova_robot} equipped with a Sake EZGripper \cite{sake_gripper} to do so.
Fig. \ref{fig:packagePacking} show different frames of this experiment.

%-----------------------------------------------------------------------

\begin{figure} %[h]
    \centering
    \includegraphics[width=0.75\linewidth]{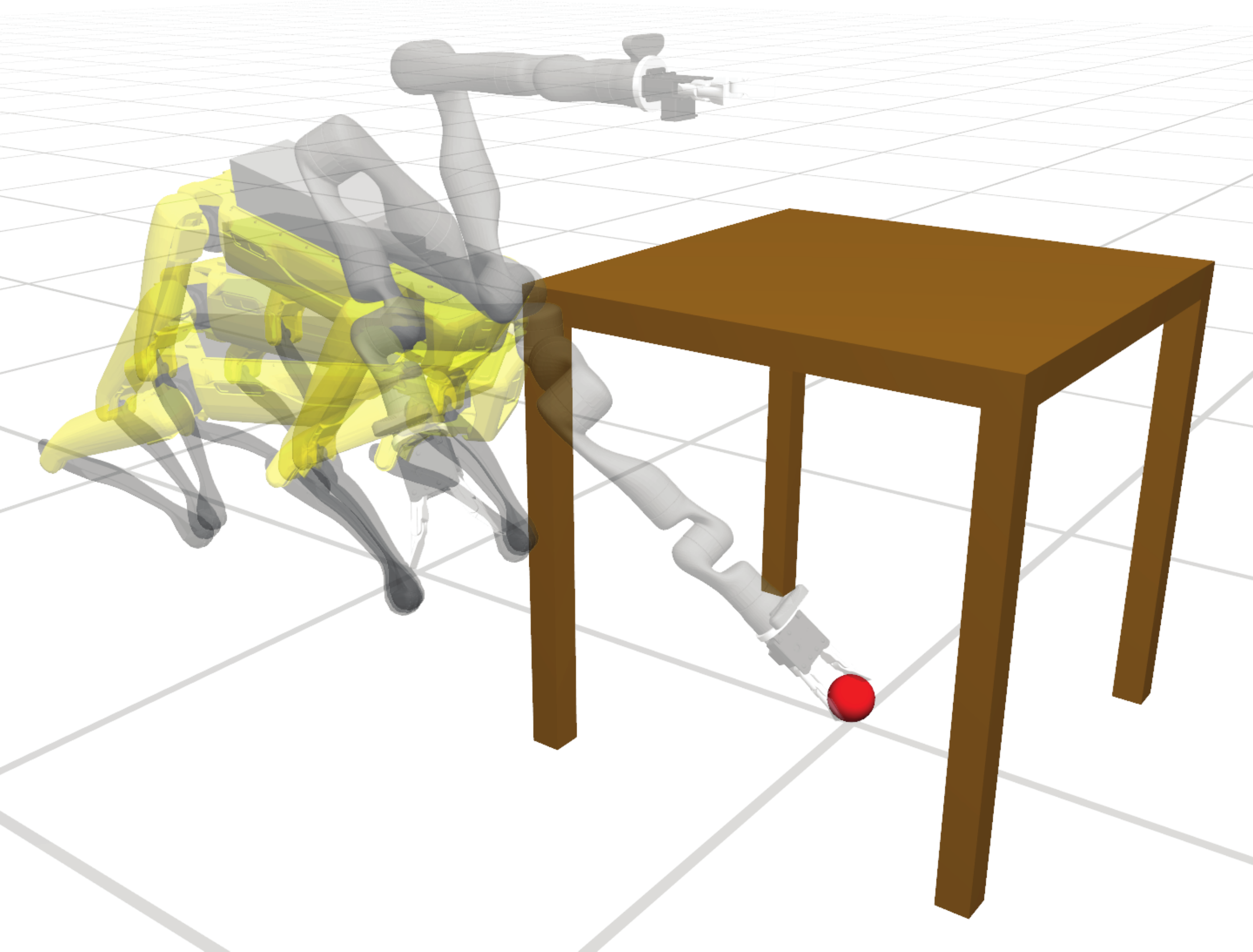}
    \caption{The combined platform consisting of Spot and Kinova are retrieving a ball from under a table. Several poses of the motion trajectory are overlaid. The motion both start and end at the most upright pose.}
    \label{fig:tableReach}
\end{figure}

\noindent \textbf{Table Reach.}
A Kinova arm mounted on a Spot robot (see \cite{zimmermann_gofetch_2021}) is tasked with retrieving an object that lies under a table.
The table is modeled by five box primitives, one for each leg, and one for the plate.
%The presented path planning framework manages to find a smooth, collision-free trajectory for the combined platform such that it swiftly picks up the object and places it onto the table.
The grasp maneuver is shown in simulation in the video, and is depicted in Fig. \ref{fig:tableReach}.

%-----------------------------------------------------------------------

\noindent \textbf{Gap Handover.}
The YuMi robot is used to hand over a package from one gripper to the other through a gap in a wall.
The wall is modeled with four rectangle collision primitives.
The video shows the experiment both in simulation and performed on the physical robot.

%-----------------------------------------------------------------------

\begin{figure} %[h]
    \centering
    \includegraphics[width=1.0\linewidth]{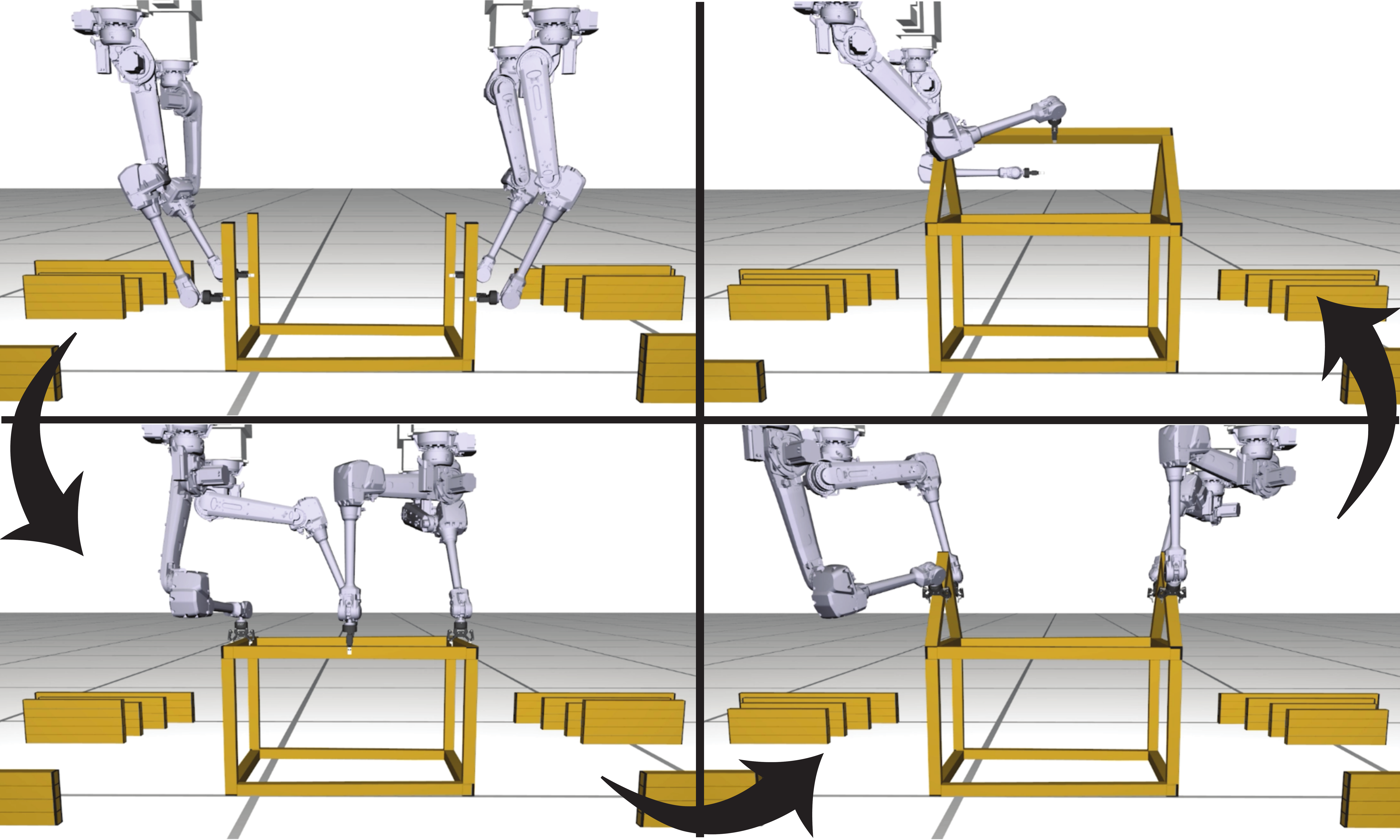}
    \caption{Four different frames of the \emph{house assembly} experiment: The RFL robots cooperatively assembly the structure of a small wooden house.}
    \label{fig:houseAssembly}
\end{figure}

\noindent \textbf{House Assembly.} 
This demonstration employs a kinematic model of the Robotic Fabrication Lab (RFL) setup at ETH Zurich \cite{rfl_setup}, which includes four large robotic arms connected by bridges.
These robot arms cooperatively assemble a simplified house by planning collision-free paths around each other as well as the structure that has already been built.
While a robot is holding a building block in its gripper, the corresponding collision primitive becomes part of the robot, and otherwise it is treated as an obstacle.
The assembly is broken down into individual pick and place tasks, where the assembly order and task assignment is done by the user.
Fig. \ref{fig:houseAssembly} shows different scenes from the assembly process.

%-----------------------------------------

\begin{table*}%[ht]
    \begin{center}
        \caption{Quantitative results for the different experiments. The state dimension is the sum of all robot state sizes in the corresponding scene.
        The last two columns show the average computation time per solver iteration and the total computation time (by multiplying the values of the two columns before), respectively.}
        \label{tab:quantitative_results}
        
        \renewcommand{\arraystretch}{1.2}
        \begin{tabular}{l | c | c | c | c | c | c | c}
            \textbf{Experiment} & \# Robots & \# Primitives & State Dim. & \# Traj. Steps &\# Iterations & $\varnothing$ Time / Iteration & Total Time \\
            \hline \hline
            Interactive Avoidance & 1 (dual arm) & 19 & 20 & 1 & 9 & 0.003 $s$ & 0.027 $s$ \\
            \hline
            Legged Armada & 8 - 10 & 8 - 12 & 48 - 60 & 100 & 446 - 1577 & 0.217 - 0.488 $s$ & 97 - 770 $s$ \\
            \hline
            Package Packing & 1 & 5 & 13 & 160 & 32 & 0.218 $s$ & 7 $s$ \\
            \hline
            Table Reach & 1 & 12 & 13 & 120 & 215 & 0.783 $s$ & 168 $s$ \\
            \hline
            Gap Handover & 1 (dual arm) & 6 & 20 & 160 & 184 & 0.373 $s$ & 69 $s$ \\
            \hline
            House Assembly & 4 & 16 - 33 & 48 & 30 & 15 - 68 & 0.004 - 0.080 $s$ & 0.06 - 5.44 $s$ \\
            \hline
        \end{tabular}
        \renewcommand{\arraystretch}{1.0}
    \end{center}
\end{table*}

%--------------------------------------------------------------------------------------------------

\subsection{Primitive Configuration}
\begin{figure} %[h]
    \centering
    \includegraphics[width=1.0\linewidth]{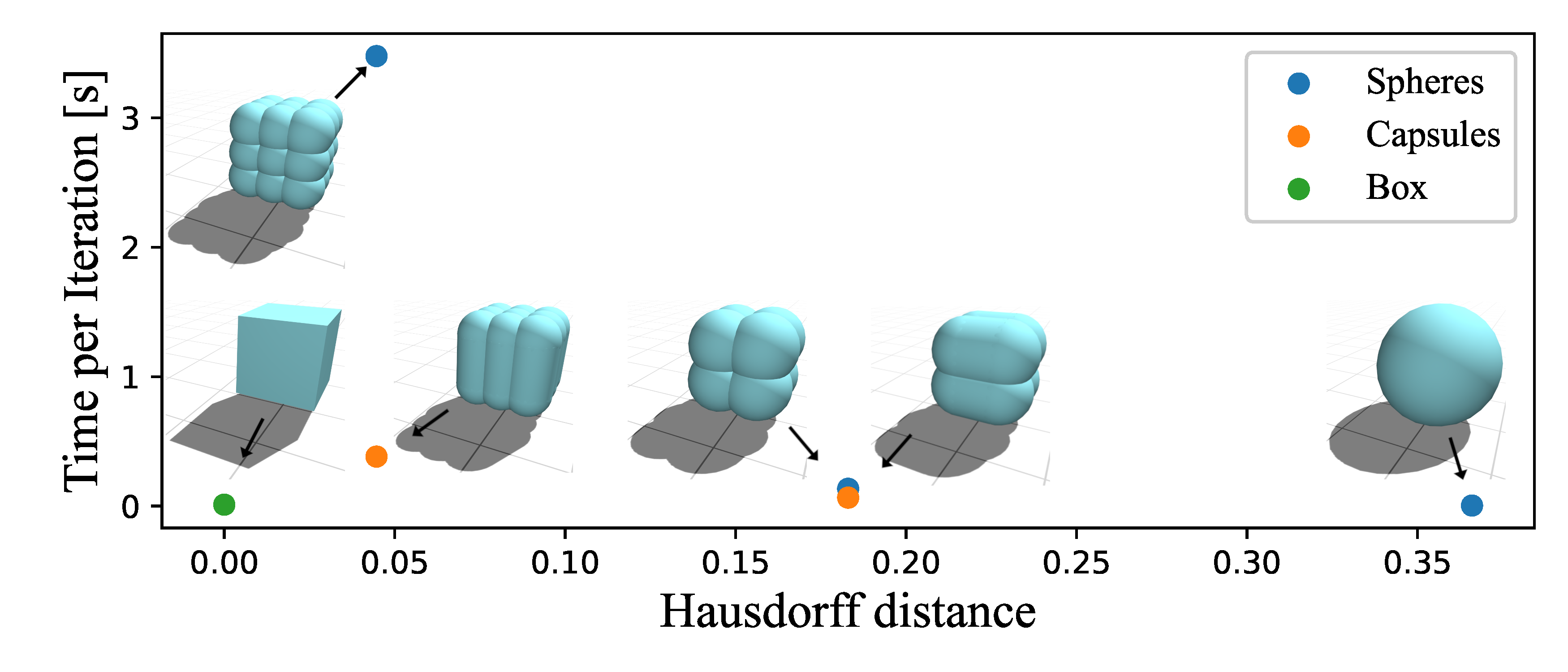}
    \caption{A free-floating rigid body and a world obstacle, both in form of a box, are approximated by different number of primitives. We compare the Hausdorff distance with the average run-time to compute Newton steps. Approximating both boxes with multiple spheres or capsules decreases the Hausdorff distance, but has a quadratic effect on the computation time. For this example, it is highly beneficial to use a box primitive.}
    \label{fig:primitive-approximation}
\end{figure}

\noindent Generally, using more primitives has the greater potential to better approximate the shape of a body.
However, more primitives lead to a higher computational cost. 
We illustrate this using a simple example shown in Fig. \ref{fig:primitive-approximation}.
There, we solve a trajectory optimization problem involving two cube-shaped obstacles.
We approximate one cube using a varying number of spheres and capsules to compare the timings.
We show that using more primitives reduces the approximation error, given here by the Hausdorff distance, but increases the computation time per iteration.
We can see that for the same Hausdorff distance, fewer capsules than spheres are needed, and despite the slight increase in distance computation time per capsule, the overall time is significantly shorter.
Naturally, using the cube itself as the primitive is the ideal choice, and this is also evident in the figure.
We additionally show in Table \ref{tab:dca-pair-times} the times required to compute the shortest distance between different primitive pairs using our method.
Again, making the primitive higher-dimensional introduces additional Newton steps, which also contributes to higher computation times, but the overall effect for the path planning problem is a reduction in time.

\begin{table}%[]
    \caption{Number of Newton steps and average time per step for the distance computation between each pair of primitives. Time measurements are averaged over 10'000 runs and expressed in microseconds.}
    \label{tab:dca-pair-times}
    \renewcommand{\arraystretch}{1.2}
    \begin{tabular}{l|c|c|c|c}
                 & Sphere     & Capsule & Rectangle & Box       \\ \hline \hline
    Sphere        & 0 / 0.0 & -             &    -       & -           \\ \hline
    Capsule & 1-2 / 0.85 & 1-3 / 1.54    & -          & -           \\ \hline
    Rectangle    & 1-7 / 1.65 & 1-5 / 2.30    & 1-11 / 3.34 &  -          \\ \hline
    Box          & 1-5 / 2.25 & 1-11 / 3.13    & 2-9 / 3.86 & 2-14 / 4.60 \\ \hline
    \end{tabular}
    \renewcommand{\arraystretch}{1.0}
\end{table}

\subsection{Comparison to Other Motion Planners}
\label{subsec:comparisonOtherPlanners}

\begin{figure} %[h]
    \centering
    \includegraphics[width=1.0\linewidth]{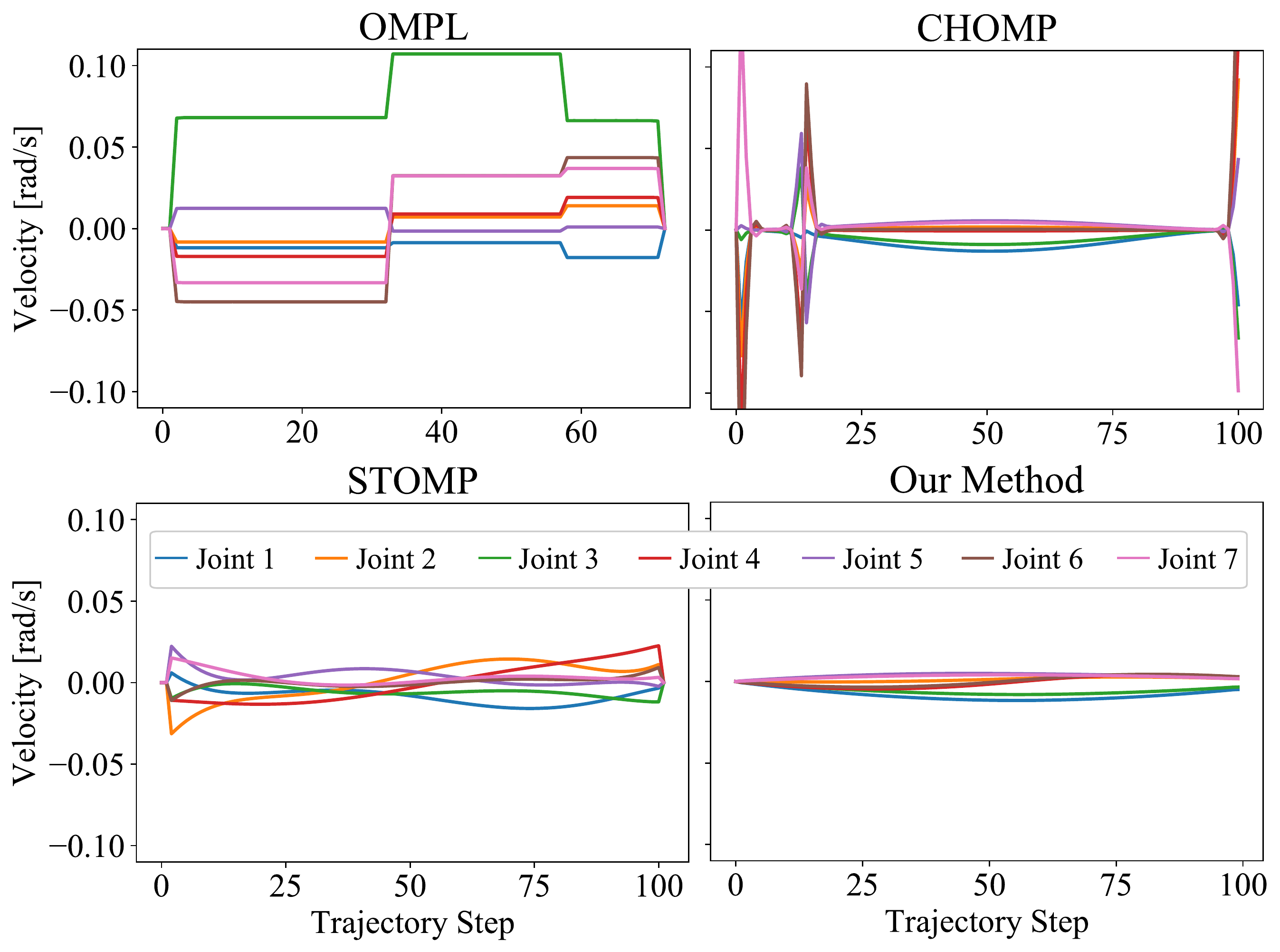}
    \caption{Resulting joint velocities for different motion planners for the experiment described in section \ref{subsec:comparisonOtherPlanners} using the largest box obstacle. The number of trajectory steps was chosen by the individual planners, and we chose 100 steps for our method for all experiments. Our method produced smoother velocities and found a more direct path.}
    \label{fig:chomp-stomp}
\end{figure}

\noindent We conducted several experiments in order to compare our method to three well-known motion planners: OMPL \cite{sucan_ompl_2012}, CHOMP \cite{chomp}, and STOMP \cite{stomp}.
Hereby, we used the default parameter settings (including initialization procedures) set in their individual ROS implementations.
The experimental setup involved a Kinova arm \cite{kinova_robot} tasked with finding a collision-free path around an obstacle modeled by either a sphere, a capsule (a cylinder in ROS), or a box.
We ran the experiment multiple times, where we gradually increased the size of the individual obstacles to make the path planning problem more elaborate.
All methods usually succeeded in finding a feasible path around the obstacle.
An overview of the individual measurements is given in Table \ref{tab:comparison_other_planners}.
We note that the overall computation time is dependent on the convergence criteria of the individual planners.
When using the default settings, the computation times of all methods were usually comparable.
However, we noticed a difference in the quality of the solution.
Especially for larger obstacles, some of the planners seemed to struggle to find a smooth path.
Our method usually provided a smoother and more direct path, both in terms of joint angles (see Fig. \ref{fig:chomp-stomp}) as well as for the robot's end-effector (see Table \ref{tab:comparison_other_planners}).

\begin{table}%[ht]
    \caption{Performance comparison to other planners. The table shows both the total computation time (in seconds) as well as the distance travelled by the robot's end-effector (in meters) for the final solution. All measurements have been averaged over several runs using the largest obstacle sizes. While the computation times are usually comparable, our method typically provides a smoother and more direct path.}
    \label{tab:comparison_other_planners}

    \centering    
    \renewcommand{\arraystretch}{1.2}
    \begin{tabular}{l  c | c | c | c | c }
         & \textbf{Planner} & OMPL & CHOMP & STOMP & Ours \\
        \hline \hline
        \multirow{2}{*}{Sphere} & Time [s] & 0.258 & 2.211 & 4.163 & 0.623 \\
        \cline{2-6}
         & Dist. [m] & 3.601 & 1.316  & 1.426  & 0.920 \\
        \hline
        Capsule / & Time [s] & 0.371 & 2.31 & 2.95  & 0.652 \\
        \cline{2-6}
        Cylinder & Dist. [m] & 1.491 & 1.615 & 1.449 & 0.972 \\
        \hline
        \multirow{2}{*}{Box} & Time [s] & 0.158 & 1.830 & 0.348 & 0.822 \\
        \cline{2-6}
         & Dist. [m] & 1.343 & 1.166 & 1.200 & 0.740 \\
        \hline
    \end{tabular}
    \renewcommand{\arraystretch}{1.0}
\end{table}
\section{Discussion}
%We presented a generic trajectory optimization method that can generate collision-free, smooth motions for multiple robots and dynamic obstacles.
%Thereby, we formulate the distance computation between different collision primitives as a small optimization problem and show how to compute its derivatives.
%\noindent To summarize, our approach provides a unified, straight-forward framework that can be applied to various collision primitives and can safely handle numerical issues that can arise when computing distances and its derivatives.
%The experiments shown demonstrate the efficacy of our method, in simulation and the real-world, testing self-collision, robot-collision, and world-collision avoidance.
%In addition, it outperforms previous methods in terms of quality, and is competitive in terms of timing.

\noindent To summarize, our approach provides a unified, straight-forward framework that can be applied to various collision primitives, and safely handles numerical issues that can arise when computing distances and its derivatives.
Therefore, our distance computation scheme can seamlessly be integrated into other path planners that profit from these properties.
In terms of limitations, our overall trajectory optimization framework suffers from the same drawbacks as other gradient-based methods: it can only find a local minimum.
Consequently, in very cluttered environments and tight spaces, our method might fail to find a collision-free trajectory without a proper initialization.
One solution would be to find an initial trajectory using a sampling-based method and use it as a starting point for the optimization.
However, due to the relatively quick iteration time and predictable behaviour, we believe it to be simpler for a user to provide input and guide the process toward a feasible solution.
Additionally, we experimented with an automatic continuation method that shows promising results.
With these methods, infeasible cases can be resolved by starting the optimization in an obstacle-free environment, such that an initial, collision-free trajectory can be generated.
Then, obstacles are gradually introduced into the scene while keeping the optimization process running.
Further analysis in this regard will be part of future investigations.
Another avenue to explore is the automatic generation of primitives that would cover the objects.
Currently, this is done manually based on previous experience.
However, we believe that finding an optimal configuration of primitives that achieves the best performance would be of great value.

%Second, for some collision primitives, it is important to start from an initial collision-free path, as our approach is not always able to recover in case a collision along the trajectory already occurred.
%For example, this can be the case when a box primitive is used, as the parameters $\t$ that are computed for the shortest distance may correspond to a point that lies \emph{within} the box.
%Therefore, when two boxes have already collided, the distance computation and its derivatives result in zero, which is why the framework is unable to "push" the primitives apart.
%One idea of how to resolve this issue would be to add additional constraints to the distance computation, such that the resulting point always lies on the primitive's surface.
%This will be further investigated in future work.

%%% Bibliography
\bibliographystyle{IEEEtran}
\bibliography{references}

\end{document}